\RequirePackage{fix-cm}
\documentclass[smallextended]{svjour3}       
\smartqed  
\usepackage{graphicx}
\usepackage{breakcites}

\begin{document}

\title{Synthesis of a Six-Bar Gripper Mechanism for Aerial Grasping
}

\titlerunning{Synthesis of a Six-Bar Gripper Mechanism for Aerial Grasping}        

\author{Rajashekhar V.S \and Rokesh Laishram \and Kaushik Das \and Debasish Ghose 
}


\institute{Rajashekhar V.S \at
              Department of Aerospace Engineering, Indian Institute of Science, Bangalore 560012  \\
              \email{vsrajashekhar@gmail.com}           
           \and
           Rokesh Laishram \at
             Robert Bosch Centre for Cyber-Physical Systems, Indian Institute of Science, Bangalore 560012
             \email{laishrams@iisc.ac.in}  
                       \and
           Kaushik Das \at
             Tata Consultancy Services, Research and Innovation Lab, Bangalore  \\
             \email{kaushik.da@tcs.com}  
          \and
           Debasish Ghose \at
             Department of Aerospace Engineering, Indian Institute of Science, Bangalore 560012 \\
             \email{dghose@iisc.ac.in}       
}


\maketitle

\begin{abstract}
In this paper, a 1-DoF  gripper mechanism has been synthesized for the type of mechanism, number of links and joints, and the dimensions of length, width and thickness of links. The type synthesis is done by selecting the proper class of mechanism from Reuleaux's six classes of mechanisms. The number synthesis is done by using an algebric method. The dimensions of the linkages are found using the geometric programming method. The gripper is then modelled in a computer aided design software and then fabricated using an additive manufacturing technique. Finally the gripper mechanism with DC motor as an actuator is mounted on an Unmanned Aerial Vehicle (UAV) to grip a spherical object moving in space. This work is related to a task in challenge 1 of Mohamed Bin Zayed International Robotics Challenge (MBZIRC)-2020.  
\keywords{Aerial manipulation \and Unmanned Aerial Vehicles \and Mechanism Synthesis \and Dimensional Synthesis \and Geometric Programming}
\end{abstract}

\section{Introduction}
Usage of unmanned aerial vehicles (UAV) have increased in the recent past for commercial and rescue operations. During these operations, the UAVs have to transport or manipulate objects of interest. A gripper attached to the UAV would aid in this operation. The manipulating devices are usually attached to multi-rotor UAVs. The manipulating devices that are normally attached to UAVs include grippers, manipulators, and cable and tether \cite{ding2018review}. The grippers are special kind of manipulators that are used to grasp objects and release them at the target location. Although the range of grippers available for aerial use are not many, their stability is high when compared to standard manipulators and cables \cite{xilun2019review}. \\
In the literature, there has been several mention of grippers being used in UAV drones \cite{mellinger2011design} and helicopters \cite{bejar2019helicopter} for aerial grasping. They include suction grippers \cite{kessens2016versatile}, magnetic grippers \cite{fiaz2017passive}, electro-permanent magnetic gripper \cite{gawel2017aerial} and multi-finger gripper \cite{vazquez2015optimal}. In the work shown in \cite{zhao2017whole}, the whole body is used for aerial manipulation by using a transformable multirotor with 2D multi-links. A literature survey on the aerial manipulation \cite{ruggiero2018aerial} shows that the following two solutions are commonly adopted to accomplish the grasping task (1) The gripper or a multi-fingered hand is directly mounted on the UAV, or  (2) A robotic arm, with the gripper as the end-effector, is attached to the UAV. The gripper synthesised in this paper has multi-fingers actuated by a multi-loop mechanism based on solution (1).  \\
In this paper, a new 1-DoF gripper mechanism, specifically designed for use with UAVs, has been synthesized for type, number of joints and links, and dimensions. Section \ref{subsec:conditionforsyn} lists the conditions for synthesis. Section \ref{subsec:typesyn} explains the type synthesis of the mechanism. In Section \ref{subsec:numbersyn} the number synthesis is carried out to find the number of links and joints needed for the mechanism. Section \ref{subsec:dimsyn} derives the dimensions of the links such as the length, width and thickness. The experimental results are presented in Section \ref{sec_exp} before the concluding remarks given in Section \ref{sec_conclusion}.       
\section{Synthesis of the Mechanism}
\label{sec_synthesis}
\subsection{Conditions for Synthesis}
\label{subsec:conditionforsyn}
The gripper is to be mounted on a quadrotor platform which has a maximum take-off mass of 13 kg (with 3.5 kg payload) and a hovering time of about 12 minutes. The quadrotor platform should have a good motion capture system. A 1-DoF gripper mechanism is to be used. The number of links used in the mechanism should be as low as possible. The gripper is required to grab a spherical object that is 10 cm to 15 cm in diameter. The gripper will be subjected to vibrations due to the high speed of the propellers.  
\subsection{Type Synthesis}
\label{subsec:typesyn}
The type of mechanism to be used is chosen by taking into consideration the Reuleaux's six classes of mechanisms \cite{moon2007machines}. The six elements are crank chain, pulley chain, screw chain, ratchet chain, gear chain and cam chain. The crank chain produces a lightweight and reliable mechanism. The pulley chain requires a heavy setup. The screw chain would yield a slow and heavy mechanism. The ratchet chain enables motion in only one direction. The gear chain is heavy and requires a complex design. The cam chain is used for high-speed applications. Among the six types, the crank chain is chosen because it would yield a reliable and lighter mechanism.   
\subsection{Number Synthesis}
\label{subsec:numbersyn}
It is done as mentioned in \cite{norton2011kinematics}. The odd number of linkages require an even number of actuators and vice-versa. Starting with $4$ links, although a manipulator can be made \cite{mermertacs2004optimal}, it is not possible to produce a gripper with a four-bar mechanism that can capture a spherical object. If $5$ links are used, then the gripping mechanism has 2-DoF. Using $6$ links, we substitute the values in the number synthesis given below \cite{norton2011kinematics}.
\begin{equation}
\label{equ:numbersynthesis}
L-3-M=T+2Q+3P+4H
\end{equation}     
where, $L$=Number of links, $M$= Degrees of freedom, $T$= Number of ternary links, $Q$= Number of quaternary links, $P$= Number of pentagonary links and $H$= Number of hexagonary links. \\
Based on the given conditions, $L=6$ and $M=1$. We get $T+2Q+3P+4H=2$. Therefore, $P=H=0$ and $T=0,1,2$, $Q=0,1$.\\
The total number of links is given by, 
\begin{equation}
\label{equ:numbersynthesistotalequations}
L=B+T+Q+P+H
\end{equation} 
Substituting the above values in (\ref{equ:numbersynthesistotalequations}), we get\\
$Case$ $1:$ $B=4$ and $T=2$ \\
$Case$ $2:$ $B=5$ and $Q=1$ \\
This means that the mechanism with six links can have four binary links and two ternary links or five binary links and one quaternary link.
\subsubsection{Designing the Mechanism}
\begin{figure}[h] 
\centering
\includegraphics[width=0.7\textwidth]{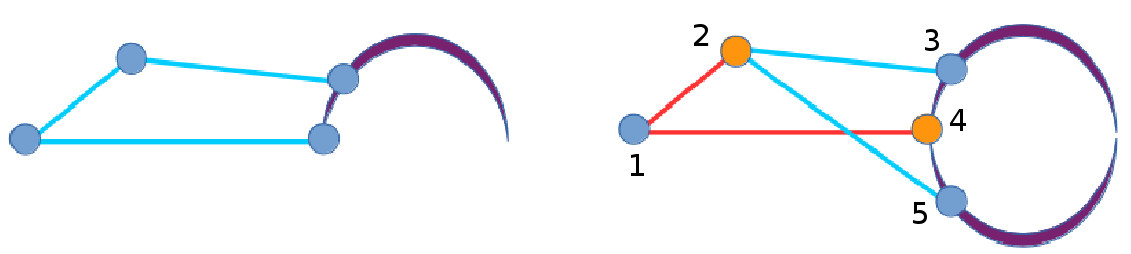}
\caption{Development of the gripper mechanism (a) The four-bar mechanism formed using the four binary links (b) The two binary links are added to the mechanism and the six bar mechanism is formed using four binary links and two ternary links}
\label{fig_gripperdevelopment}
\end{figure}
\begin{figure}[h] 
\centering
\includegraphics[width=0.8\textwidth]{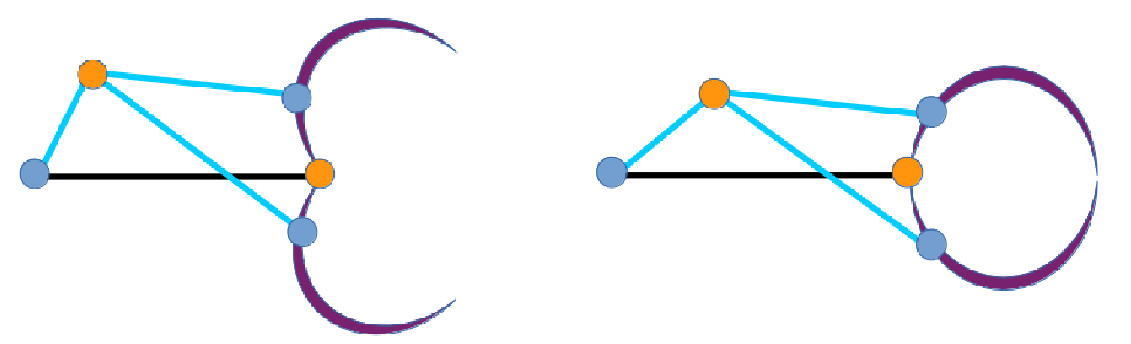}
\caption{The synthesized gripper mechanism (a) open position (b) close position}
\label{fig_conceptgripper}
\end{figure}
$Case$ $1$ of the number synthesis done above yields the type of gripper as shown in Figure \ref{fig_gripperdevelopment} after following the steps in \cite{erdman1986type} and isomorphism steps mentioned in \cite{tsai2000mechanism}. Figure \ref{fig_conceptgripper} (a) shows the gripper in an open position. Figure \ref{fig_conceptgripper} (b) shows the gripper in closed position. In the mechanism, there are totally six links and seven joints. Of the six links, four are binary and two are ternary. The joints are binary or ternary in nature. The joints 1,3,5 depict a single revolute joint and joints 2,4 depict two revolute joints at that location. The line joining the joints 1 and 4 is the fixed link and the rest are moving links. When the joint 1 rotates clockwise and anti-clockwise, the end effector closes and opens respectively.\\
The $Case$ $2$ of the number synthesis done above does not yield a gripper mechanism after following the steps in \cite{erdman1986type} and isomorphism steps mentioned in \cite{tsai2000mechanism}. Hence $Case$ $1$ is taken into consideration for designing the gripper.
\subsection{Dimensional Synthesis}
\label{subsec:dimsyn}
The ratio of the dimension of the linkages that are used in the mechanism is found here. These cross-sectional parameters have an effect on stress and inertial force.      
\subsubsection{Estimating the Lengths of the Links}
\label{subsub_length}
\begin{figure} [h]
\centering
\includegraphics[width=0.6\textwidth]{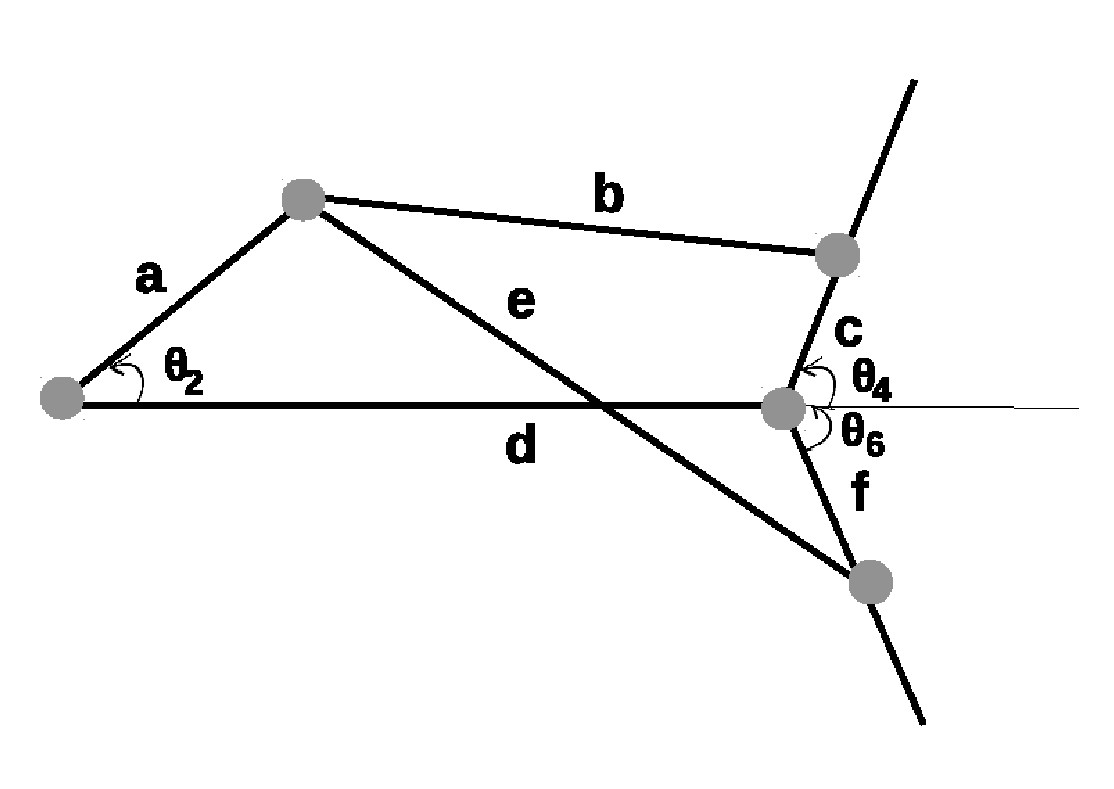}
\caption{The notations used in the dimensional synthesis}
\label{fig_gripperdimsyn}
\end{figure}
\begin{figure} [h]
\centering
\includegraphics[width=0.5\textwidth]{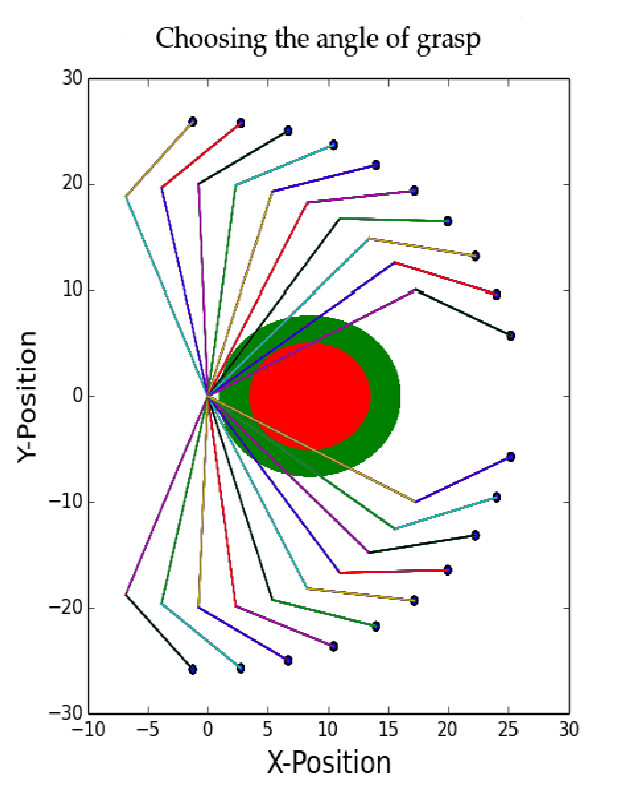}
\caption{Choosing the range of angles as given in Table 1}
\label{fig_choosingtheangle}
\end{figure}

The dimensional synthesis is done by geometric programming as mentioned in \cite{rao1979synthesis}. It is briefly explained here with a numerical example. The two four-bar mechanisms with link dimensions $a,b,c,d$ and $a,e,f,d$, and input angle $\theta_{2}$ and output angles $\theta_{4}$ and $\theta_{6}$ are as shown in Figure \ref{fig_gripperdimsyn}. The two displacement equations are as follows:
\begin{equation}
2ad\cos\theta_{2} - 2cd\cos\theta_{4} + (a^2 - b^2 + c^2 + d^2) - 2ac \cos(\theta_{2}-\theta_{4}) = 0
\label{equ:disp1}
\end{equation} 
\begin{equation}
2ad\cos\theta_{2} - 2fd\cos\theta_{6} + (a^2 - e^2 + f^2 + d^2) - 2af \cos(\theta_{2}-\theta_{6}) = 0
\label{equ:disp2}
\end{equation} 
The actual output angle $\theta_{4i}$ and $\theta_{6i}$ for a particular input angle $\theta_{2i}$ includes some error $\epsilon_{1i}$ and $\epsilon_{2i}$ when compared with the desired value $\theta_{4di}$ and $\theta_{6di}$. Hence,
\begin{equation}
\theta_{4i} = \theta_{4di}+\epsilon_{1i}
\label{equ:struct1}
\end{equation}
\begin{equation}
\theta_{6i} = \theta_{6di}+\epsilon_{2i}
\label{equ:struct2}
\end{equation}
where, $\epsilon_{1i}$ and $\epsilon_{2i}$ are structural errors at the given $\theta_{2i}$. Substituting (\ref{equ:struct1}) and (\ref{equ:struct2}) into  (\ref{equ:disp1}) and (\ref{equ:disp2}), respectively, and assuming $\sin \epsilon \approx \epsilon$ and $\cos \epsilon \approx 1$ for small values of $\epsilon$, we get 
\begin{equation}
\label{equ:error1}
\epsilon_{1i}=\frac{K + 2ad\cos \theta_{2i} - 2cd\cos \theta_{2di} - 2ac\cos \theta_{2i} \cos(\theta_{4di}-\theta_{2i})}{-2ac\sin(\theta_{4di}-\theta_{2i}) - 2cd \sin \theta_{4di}}
\end{equation}
\begin{equation}
\label{equ:error2}
\epsilon_{2i}=\frac{L + 2ad\cos \theta_{2i} - 2fd\cos \theta_{2di} - 2af\cos \theta_{2i} \cos(\theta_{6di}-\theta_{2i})}{-2af\sin(\theta_{6di}-\theta_{2i}) - 2fd \sin \theta_{6di}}
\end{equation}
where
\begin{equation}
\label{equ:k}
K = a^2 -b^2 +c^2 +d^2 
\end{equation} 
and 
\begin{equation}
\label{equ:l}
L = a^2 -e^2 +f^2 +d^2 
\end{equation}
At some values of $\theta_{2}$, the values of $\theta_{4}$ and $\theta_{6}$ which is generated by the mechanism is to be made equal to the desired values of $\theta_{4d}$ and $\theta_{6d}$, respectively which are known as precision points. The errors $\epsilon_{1i}$ and $\epsilon_{2i}$ are minimized when the functions $F_1$ and $F_2$ are minimized. Hence, the objective function for the minimization problem is taken to be the sum of the squares of $\epsilon$ at a number of precision points. 
\begin{equation}
F_1= \sum_{i=1}^{n} \epsilon_{1i}^2
\end{equation}
 \begin{equation}
F_2= \sum_{i=1}^{n} \epsilon_{2i}^2
\end{equation}
where, $n$ is the number of precision points. \\
For simplification, let $a \ll d$ and the value of $\epsilon_0=0$ at $\theta_{2i} = \theta_{20}$. Substituting these in (\ref{equ:error1}) and (\ref{equ:error2}), the following are obtained:
\begin{equation}
\label{equ:K}
K=2cd\cos \theta_{4di} + 2ac\cos \theta_{20}\cos(\theta_{4d0}-\theta_{20}) - 2ad\cos \theta_{20}
\end{equation}
\begin{equation}
\label{equ:L}
L=2fd\cos \theta_{6di} + 2af\cos \theta_{20}\cos(\theta_{6d0}-\theta_{20}) - 2ad\cos \theta_{20}
\end{equation}
Based on the space available below the drone, we set the following constraint.
\begin{equation}
\label{equ:adcondition}
\frac{3a}{d}\leq1
\end{equation}
Thus objective functions are written as follows.\\
\begin{equation}
\label{equ:objcondition}
F_{1} = \sum_{i=1}^{n} \frac{a^{2}(\cos \theta_{2i} - \cos \theta_{20})^2 - 2ac(\cos \theta_{2i} - \cos \theta_{20})(\cos \theta_{4di} - \cos \theta_{4d0}) }{c^{2}\sin^{2}\theta_{4di}}
\end{equation}
\begin{equation}
F_{2} = \sum_{i=1}^{n} \frac{a^{2}(\cos \theta_{2i} - \cos \theta_{20})^2 - 2af(\cos \theta_{2i} - \cos \theta_{20})(\cos \theta_{4di} - \cos \theta_{4d0}) }{f^{2}\sin^{2}\theta_{4di}}
\end{equation}
Considering the loop 1 of the four bar mechanism, the normality and orthogonality conditions are as follows:
\begin{equation}
\label{equ:ortho}
\Delta_{1}^{*}+\Delta_{2}^{*}=1
\end{equation}
\begin{equation}
\label{equ:nor1}
2\Delta_{1}^{*}+\Delta_{2}^{*}=0
\end{equation}
\begin{equation}
\label{equ:nor2}
2\Delta_{1}^{*}+0.5\Delta_{2}^{*}+\Delta_{3}^{*}=0
\end{equation}
Solving (\ref{equ:ortho}), (\ref{equ:nor1}) and (\ref{equ:nor2}), we get $\Delta_{1}^{*}=-1$,$\Delta_{2}^{*}=2$ and $\Delta_{3}^{*}=1$. The equation of the dual function for maximum value is given by 
\begin{equation}
\label{equ:dualfunction}
\textit{v}(\Delta^{*})= \left(\frac{c_1}{\Delta_{1}^{*}}\right)^{\Delta_{1}^{*}} \left(\frac{c_2}{\Delta_{2}^{*}}\right)^{\Delta_{2}^{*}} \left(\frac{c_3}{\Delta_{3}^{*}}\right)^{\Delta_{3}^{*}}
\end{equation} 
where, $c_1$, $c_2$ and $c_3$ are the coefficients of posynomial terms in (\ref{equ:adcondition}) and (\ref{equ:objcondition}). \\
Figure \ref{fig_choosingtheangle} shows how the values of the angles were chosen in Table \ref{table_angles}. The lines from the origin are at an angle of $38^{\circ}$ to $110^{\circ}$. It can be observed that the smallest spherical object can be captured at $45^{\circ}$  and the largest spherical object can be captured at $70^{\circ}$.
The link lengths of the four-bar mechanism are calculated as follows. \\
\begin{table}[h]
\caption{Choosing the precision points}
\begin{center}
\begin{tabular}{|c|c|c|c|c|c|c|c|}
 \hline 
 Precision Point, i & 1 & 2 & 3 & 4 & 5 & 6 & 7 \\ 
 \hline 
 Input $\theta_2$ & 0 & 5 & 10 & 15 & 20 & 25 & 30 \\ 
 \hline 
 Output $\theta_4$ & 135 & 120 & 105 & 90 & 75  & 60 & 45 \\ 
 \hline  
 \end{tabular} 
 \end{center} 
    \label{table_angles}    
\end{table}
The structural error is zero at precision point 4 ($\theta_{20}=15$ and $\theta_{4d0}=90$). Substituting the values in the table in (\ref{equ:objcondition}), we get 
\begin{equation}
f=0.029362689\left(\frac{a^2}{c^2}\right)+0.523816599\left(\frac{a}{c}\right)
\end{equation}
subject to $\frac{3a}{d}\leq1$.\\
We get $c_{1}=0.029362689$, $c_{2}=0.523816599$ and $c_{3}=\frac{3}{d}$. Substituting these in (\ref{equ:dualfunction}) gives,\\
$\textit{v}(\Delta)= \left(\frac{-7.0081}{d}\right)$\\
Now the individual parameters are determined as in \cite{rao1979synthesis} to be $a^*=1$, $c^*=0.1122$ and $d^*=3$. Substituting these in (\ref{equ:k}) and (\ref{equ:K}), we get $b^*=3.9689$.\\
Thus, the optimal lengths of links in the first loop are $a^*=1$, $b^*=3.9689$, $c^*=0.1122$ and $d^*=3$. Similarly, it can be found for the second loop of the mechanism as $a^*=1$, $e^*=3.9689$, $f^*=0.1122$ and $d^*=3$. 
\subsubsection{Estimating the Width of the Links}
\label{subsub_width}
The change in the width of the linkages have a very little effect on the natural frequency of a mechanism \cite{yu1996effect}. Hence it can be chosen based on the diameter of the pin joint. The holes on the linkages that are used for inserting the revolute joints is taken to be $8$ mm in diameter. According to the data in \cite{lingaiah2003machine} for a hole in a plate, the width is chosen to be $26.66$ mm for the linkages $a,b,c,d,e,f$.
\subsubsection{Estimating the Thickness of the Links}
\label{subsub_thick}
\begin{figure} [h]
\centering
\includegraphics[width=0.8\textwidth]{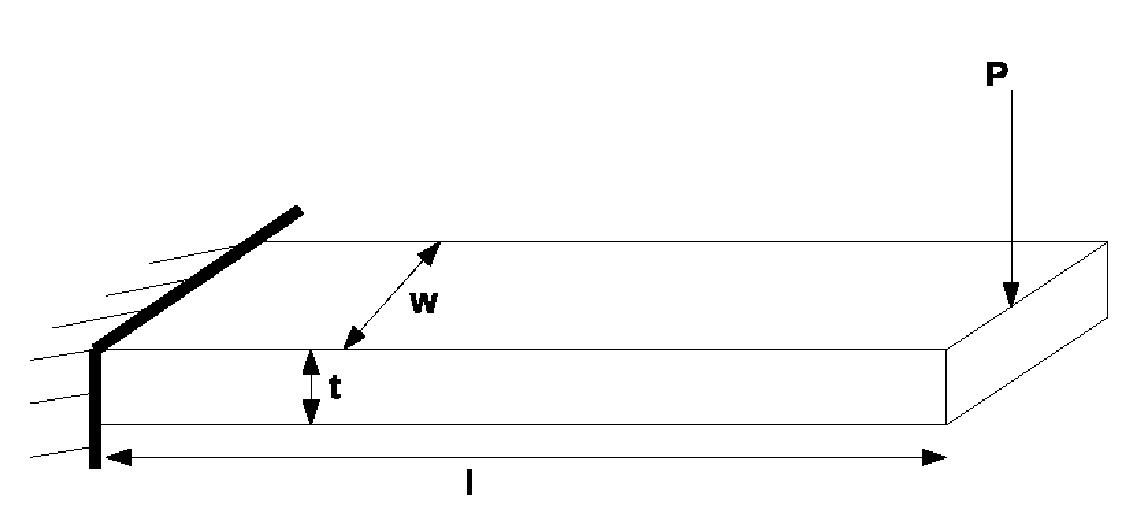}
\caption{The parameters of the link which is considered as the cantilever beam}
\label{fig_thickness}
\end{figure}
The thickness of the linkages have a greater effect on the natural frequency of a mechanism \cite{yu1996effect}. The thickness of the links are found using geometric programming as done in \cite{rao2009engineering}. The linkages are treated as a cantilever beam as shown in Figure \ref{fig_thickness}. This approach of considering the linkages as a cantilever beam is standard in the literature \cite{ahmadian2017design,yu1996effect}. The thickness is considered as a design variable and the objective function is given by:
\begin{equation}
f(X)=\rho l t w
\end{equation}  
where, $\rho$ is the density of the material, $l$ is the length of the link, $t$ is the thickness of the link and $w$ is the width of the link. The maximum stress at the fixed end is given by 
\begin{equation}
\label{equ_sigmathick}
\sigma=\frac{Mc}{I}=P l \left(\frac{t}{2}\right) \left(\frac{1}{\frac{w t^{3}}{12}}\right) = \frac{6Pl}{w t^{2}}
\end{equation} 
where, $\sigma$ is the maximum bending stress and $P$ is the load applied at the other end of the cantilever beam. Thus, the constraint from (\ref{equ_sigmathick}) becomes,
\begin{equation}
\frac{6Pl}{\sigma w t^{2}} \leq 1
\end{equation}
The geometric programming algorithm which is used to solve the above optimization condition is as follows. The ratio of lengths of linkages derived in Section \ref{subsub_length} are multiplied by a factor of 10 which is based on the area available below the drone. The values of the variables are $P=4$ N, $\rho = 1430$ kg/m$^{3}$, $l=0.1$ m and $w=0.026$ m. 
Now the function becomes, 
\begin{equation}
\label{equ_thicknessinitial}
f(X)= \left(\frac{3.718 t}{\Delta_{1}}\right)^{\Delta_{1}} \left(\frac{9.05 t^{-2} 10^{-7}}{\Delta_{2}}\right)^{\Delta_{2}}
\end{equation}
The orthogonality and normality conditions are as follows:
\begin{equation}
\label{equ_thickd1}
\Delta_{1} - 2 \Delta_{2} = 0 
\end{equation}
\begin{equation}
\label{equ_thickd2}
\Delta_{1} +  \Delta_{2} = 1 
\end{equation}
Solving (\ref{equ_thickd1}) and (\ref{equ_thickd2}), we get the values of $\Delta_{1}= \frac{2}{3}$ and $\Delta_{2}= \frac{1}{3}$. Substituting the values in   (\ref{equ_thicknessinitial}) and equating both terms, we get
\begin{equation}
\left(\frac{3.718t}{\frac{2}{3}}\right)^{\frac{2}{3}}= \left(\frac{9.05 t^{-2} 10^{-7}}{\frac{1}{3}}\right)^{\frac{1}{3}}
\end{equation}
Solving the equation, we get $t=27.47$ mm for the length of $a=100$ mm. Repeating the above steps, we get $t=15.84$ mm for the length of $b=396$ mm, $t=12.58$ mm for the length of $c=20$ mm and $t=17.70$ mm for the length of $d=300$ mm.
\section{Experimental Results}
\label{sec_exp}
\begin{figure}[h]
\centering
\includegraphics[width=0.8\textwidth]{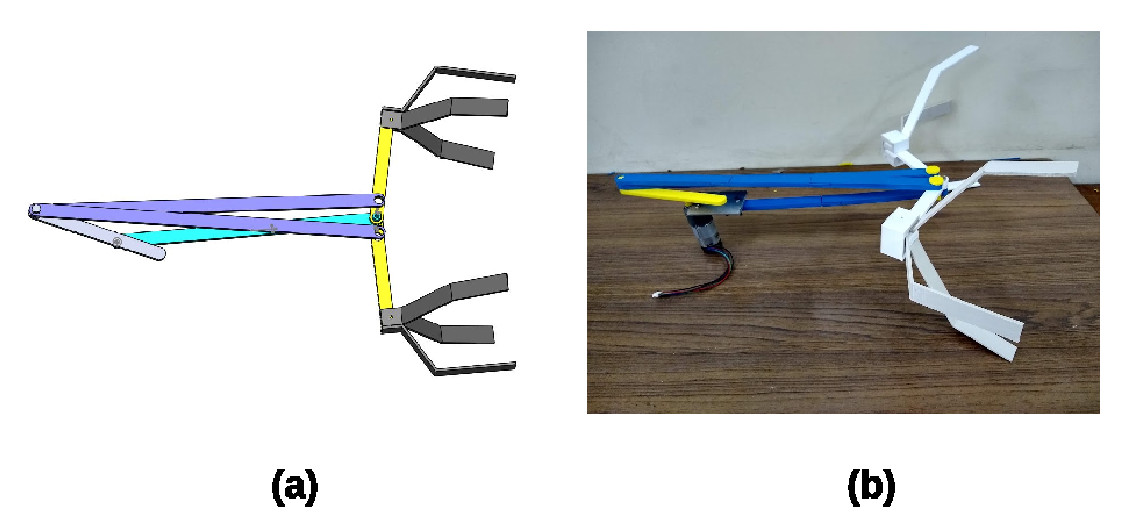}
\caption{The aerial gripper mechanism (a) The computer-aided design of the gripper (b) The prototype of the gripper with synthesised dimensions.}
\label{fig_prototype}
\end{figure}

\begin{figure}[h]
\centering
\includegraphics[width=1\textwidth]{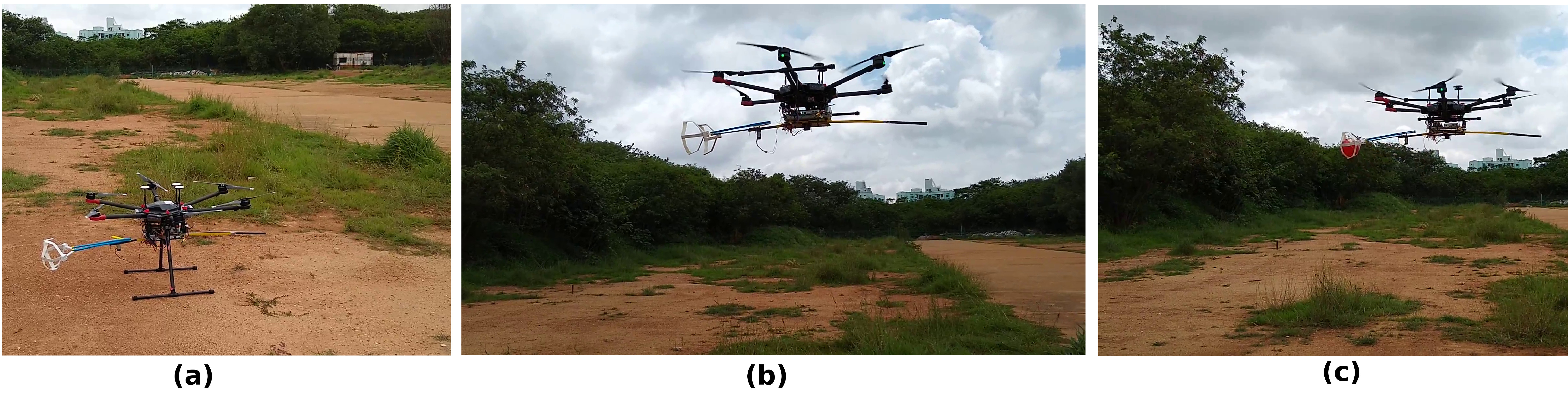}
\caption{The gripper mechanism mounted on the UAV and tested without and with the payload. (a) The gripper mechanism mounted on the UAV (b) The UAV in flight without payload (c) The UAV in flight with a payload}
\label{fig_exp}
\end{figure}

The aerial gripper mechanism was modelled in a computer aided design (CAD) software using the synthesised dimensions. It is as shown in Figure \ref{fig_prototype}(a). A prototype was made using 3D printed parts. The parts were assembled and the mechanism was tested using a DC motor with encoder. The prototype as shown in Figure \ref{fig_prototype}(b) was mounted on a UAV. The DC motor with encoder was controlled using an \textit{Arduino UNO} which received the open and close commands from the \textit{NVIDIA Jetson TX2 Development board} that was mounted on the drone. This board received the commands from the Robot Operating System (ROS) installed on a off-board computer. A motor driver board was used to give power supply to the DC motor. \\
The UAV with the gripper was initially operated to check the opening and closing of the gripper. It was then flown to test its efficacy with and without payload. A spherical object with a mass of about 0.15 kg and 150 mm in diameter was carried by the gripper in air. The UAV dropped the spherical object after few minutes of flight when the drop command was given through ROS. It was found that the system was stable with and without the payload. Some snapshots of the experiments conducted  are shown in Figure \ref{fig_exp}. The complete results of the experiments will be reported in a more detailed paper subsequently.  
\section{Conclusions}
\label{sec_conclusion}
In this work, a novel one degree of freedom planar aerial gripper has been synthesized for length, width and thickness using geometric programming. It was fabricated using 3D printed parts. The claw of the gripper was designed specially to grasp a ball of 150 mm in diameter. The claw design could be explored for unified grasping of various objects.   
\section*{Acknowledgments} 
The authors wish to acknowledge partial funding support from RBCCPS/IISc and Khalifa University, Abu Dhabi. The authors also thank the IISc-TCS MBZIRC team for their help in conducting the experiments.   
\bibliographystyle{spbasic}
\bibliography{mybibfile}

\begin{thebibliography}{20}
\providecommand{\natexlab}[1]{#1}
\providecommand{\url}[1]{{#1}}
\providecommand{\urlprefix}{URL }
\expandafter\ifx\csname urlstyle\endcsname\relax
  \providecommand{\doi}[1]{DOI~\discretionary{}{}{}#1}\else
  \providecommand{\doi}{DOI~\discretionary{}{}{}\begingroup
  \urlstyle{rm}\Url}\fi
\providecommand{\eprint}[2][]{\url{#2}}

\bibitem[{Ahmadian and Jafarishad(2017)}]{ahmadian2017design}
Ahmadian MT, Jafarishad H (2017) Design and analysis of a 3-link
  micro-manipulator actuated by piezoelectric layers. Mechanism and Machine
  Theory 112:43--60

\bibitem[{Bejar et~al.(2019)Bejar, Ollero, and Kondak}]{bejar2019helicopter}
Bejar M, Ollero A, Kondak K (2019) Helicopter based aerial manipulators. In:
  Aerial Robotic Manipulation, Springer, pp 35--52

\bibitem[{Ding et~al.(2018)Ding, Guo, Xu, and Yu}]{ding2018review}
Ding X, Guo P, Xu K, Yu Y (2018) A review of aerial manipulation of small-scale
  rotorcraft unmanned robotic systems. Chinese Journal of Aeronautics

\bibitem[{Erdman et~al.(1986)Erdman, Thompson, and Riley}]{erdman1986type}
Erdman AG, Thompson T, Riley DR (1986) Type selection of robot and gripper
  kinematic topology using expert systems. The International journal of
  robotics research 5(2):183--189

\bibitem[{Fiaz et~al.(2017)Fiaz, Toumi, and Shamma}]{fiaz2017passive}
Fiaz UA, Toumi N, Shamma JS (2017) Passive aerial grasping of ferrous objects.
  IFAC-PapersOnLine 50(1):10299--10304

\bibitem[{Gawel et~al.(2017)Gawel, Kamel, Novkovic, Widauer, Schindler, von
  Altishofen, Siegwart, and Nieto}]{gawel2017aerial}
Gawel A, Kamel M, Novkovic T, Widauer J, Schindler D, von Altishofen BP,
  Siegwart R, Nieto J (2017) Aerial picking and delivery of magnetic objects
  with mavs. In: 2017 IEEE International Conference on Robotics and Automation
  (ICRA), IEEE, pp 5746--5752

\bibitem[{Kessens et~al.(2016)Kessens, Thomas, Desai, and
  Kumar}]{kessens2016versatile}
Kessens CC, Thomas J, Desai JP, Kumar V (2016) Versatile aerial grasping using
  self-sealing suction. In: 2016 IEEE International Conference on Robotics and
  Automation (ICRA), IEEE, pp 3249--3254

\bibitem[{Lingaiah(2003)}]{lingaiah2003machine}
Lingaiah K (2003) Machine design databook, vol~2. McGraw-Hill

\bibitem[{Mellinger et~al.(2011)Mellinger, Lindsey, Shomin, and
  Kumar}]{mellinger2011design}
Mellinger D, Lindsey Q, Shomin M, Kumar V (2011) Design, modeling, estimation
  and control for aerial grasping and manipulation. In: 2011 IEEE/RSJ
  International Conference on Intelligent Robots and Systems, IEEE, pp
  2668--2673

\bibitem[{Mermerta{\c{s}}(2004)}]{mermertacs2004optimal}
Mermerta{\c{s}} V (2004) Optimal design of manipulator with four-bar mechanism.
  Mechanism and Machine Theory 39(5):545--554

\bibitem[{Moon(2007)}]{moon2007machines}
Moon FC (2007) The Machines of Leonardo Da Vinci and Franz Reuleaux: kinematics
  of machines from the Renaissance to the 20th Century, vol~2. Springer Science
  \& Business Media

\bibitem[{Norton(2011)}]{norton2011kinematics}
Norton RL (2011) Kinematics and dynamics of machinery. McGraw-Hill Higher
  Education

\bibitem[{Rao(1979)}]{rao1979synthesis}
Rao A (1979) Synthesis of 4-bar function-generators using geometric
  programming. Mechanism and Machine theory 14(2):141--149

\bibitem[{Rao(2009)}]{rao2009engineering}
Rao SS (2009) Engineering optimization: theory and practice. John Wiley \& Sons

\bibitem[{Ruggiero et~al.(2018)Ruggiero, Lippiello, and
  Ollero}]{ruggiero2018aerial}
Ruggiero F, Lippiello V, Ollero A (2018) Aerial manipulation: A literature
  review. IEEE Robotics and Automation Letters 3(3):1957--1964

\bibitem[{Tsai(2000)}]{tsai2000mechanism}
Tsai LW (2000) Mechanism design: enumeration of kinematic structures according
  to function. CRC press

\bibitem[{Vazquez et~al.(2015)Vazquez, Giacomini, Escareno, Rubio, and
  Sossa}]{vazquez2015optimal}
Vazquez J, Giacomini M, Escareno J, Rubio E, Sossa H (2015) Optimal grasping
  points identification for a rotational four-fingered aerogripper. In: 2015
  Workshop on Research, Education and Development of Unmanned Aerial Systems
  (RED-UAS), IEEE, pp 272--277

\bibitem[{Xilun et~al.(2019)Xilun, Pin, Kun, and Yushu}]{xilun2019review}
Xilun D, Pin G, Kun X, Yushu Y (2019) A review of aerial manipulation of
  small-scale rotorcraft unmanned robotic systems. Chinese Journal of
  Aeronautics 32(1):200--214

\bibitem[{Yu and Smith(1996)}]{yu1996effect}
Yu YQ, Smith M (1996) The effect of cross-sectional parameters on the dynamics
  of elastic mechanisms. Mechanism and machine theory 31(7):947--955

\bibitem[{Zhao et~al.(2017)Zhao, Kawasaki, Chen, Noda, Okada, and
  Inaba}]{zhao2017whole}
Zhao M, Kawasaki K, Chen X, Noda S, Okada K, Inaba M (2017) Whole-body aerial
  manipulation by transformable multirotor with two-dimensional multilinks. In:
  2017 IEEE International Conference on Robotics and Automation (ICRA), IEEE,
  pp 5175--5182

\end{thebibliography}
\end{document}